\pdfoutput=1

\documentclass[11pt]{article}

\usepackage{naacl2021}

\usepackage{times}
\usepackage{xcolor}
\usepackage{latexsym}
\usepackage{graphicx}
\usepackage{booktabs}       

\usepackage[T1]{fontenc}

\usepackage[utf8]{inputenc}

\usepackage{microtype}

\title{Language in a (Search) Box: \\ Grounding Language Learning in Real-World Human-Machine Interaction}

\author{Federico Bianchi\thanks{Corresponding author. Authors contributed equally and are listed alphabetically.}\\
  Bocconi University\\
  Milano, Italy \\
  \texttt{f.bianchi@unibocconi.it } \\\And
  Ciro Greco \\
  Coveo Labs\\
  New York, USA\\
  \texttt{cgreco@coveo.com} \\\And
  Jacopo Tagliabue \\
  Coveo Labs\\
  New York, USA\\
  \texttt{jtagliabue@coveo.com} \\
  }

\begin{document}
\maketitle
\begin{abstract}
We investigate grounded language learning through real-world data, by modelling a teacher-learner dynamics through the natural interactions occurring between users and search engines; in particular, we explore the emergence of semantic generalization from unsupervised dense representations outside of synthetic environments. A grounding domain, a denotation function and a composition function are learned from user data only. We show how the resulting semantics for noun phrases exhibits compositional properties while being fully learnable without any explicit labelling. We benchmark our grounded semantics on compositionality and zero-shot inference tasks, and we show that it provides better results and better generalizations than SOTA non-grounded models, such as word2vec and BERT.
\end{abstract}

\section{Introduction}
\label{sec:intro}

Most SOTA models in NLP are only \textit{intra-textual}. Models based on distributional semantics -- such as standard and contextual word embeddings~\citep{w2vec,elmo,bert} -- learn representations of word meaning from patterns of co-occurrence in big corpora, with no reference to extra-linguistic entities. 

While successful in a range of cases, this approach does not take into consideration two fundamental facts about language. 
The first is that language is a referential device used to refer to extra-linguistic objects. Scholarly work in psycholinguistics~\citep{Xu00wordlearning}, formal semantics~\citep{10.5555/335289} and philosophy of language~\citep{Quine1960-QUIWO} show that (at least some aspects of) linguistic meaning can be represented as a sort of mapping between linguistic and extra-linguistic entities. 
The second is that language may be learned based on its usage and that learners draw part of their generalizations from the observation of teachers’ behaviour~\cite{Tomasello2003}. These ideas have been recently explored by work in grounded language learning, showing that allowing artificial agents to access human actions providing information on language meaning has several practical and scientific advantages~\cite{Yu2018InteractiveGL,ChevalierBoisvert2019BabyAIAP}.

While most of the work in this area uses toy worlds and synthetic linguistic data, we explore grounded language learning offering an example in which unsupervised learning is combined with a language-independent grounding domain in a real-world scenario. In particular, we propose to use the interaction of users with a search engine as a setting for grounded language learning. In our setting, users produce search queries to find products on the web: queries and clicks on search results are used as a model for the teacher-learner dynamics.

We summarize the contributions of our work as follows:

\begin{enumerate}
    \item we provide a grounding domain composed of dense representations of extra-linguistic entities constructed in an unsupervised fashion from user data collected in the real world. In particular, we learn neural representations for our domain of objects leveraging prod2vec~\citep{Grbovic15}: crucially, building the grounding domain does not require any linguistic input and it is independently justified in the target domain~\citep{CoveoECNLP22}. In this setting, lexical denotation can also be learned without explicit labelling, as we use the natural interactions between the users and the search engine to learn a noisy denotation for the lexicon~\cite{coveoNAACLINDUSTRY21}. More specifically, we use ~\textit{DeepSets}~\citep{Cotter2018InterpretableSF} constructed from user behavioural signals as the extra-linguistic reference of words. 
    For instance, the denotation of the word ``shoes'' is constructed from the clicks produced by real users on products that are in fact~\textit{shoes} after having performed the query ``shoes'' in the search bar.
    Albeit domain specific, the resulting language is significantly richer than languages from agent-based models of language acquisition~\citep{Sowik2020ExploringSI,FitzgeraldTagliabue2020}, as it is based on 26k entities from the inventory of a real website.
    
    \item We show that a dense domain built through unsupervised representations can support~\textit{compositionality}. By replacing a discrete formal semantics of noun phrases~\citep{Heim-and-Kratzer} with functions learned over DeepSets, we test the generalization capability of the model on zero-shot inference: once we have learned the meaning of ``Nike shoes'', we can reliably predict the meaning of ``Adidas shorts''. In this respect, this work represents a major departure from previous work on the topic, where compositional behavior is achieved through either discrete structures built manually~\citep{Lu2018EntityawareIC, Krishna2016VisualGC}, or embeddings of such structures~\citep{Hamilton2018EmbeddingLQ}.
    \item To the best of our knowledge, no dataset of this kind (product embeddings from shopping sessions and query-level data) is publicly available. As part of this project, we release our code and a curated dataset, to broaden the scope of what researchers can do on the topic\footnote{Please refer to the project repository for additional information:~\url{https://github.com/coveooss/naacl-2021-grounded-semantics}.}.
\end{enumerate}

Methodologically, our work draws inspiration from research at the intersection between Artificial Intelligence and Cognitive Sciences: as pointed out in recent papers~\citep{Bisk2020ExperienceGL,bender-koller-2020-climbing}, extra-textual elements are crucial in advancing our comprehension of language acquisition and the notion of ``meaning''. While synthetic environments are popular ways to replicate child-like abilities~\citep{Kosoy2020ExploringEC,Hill2020GroundedLL}, our work calls the attention on real-world Information Retrieval systems as experimental settings: cooperative systems such as search engines offer new ways to study language grounding, in between the oversimplification of toy models and the daunting task of providing a general account of the semantics of a natural language. The chosen IR domain is rich enough to provide a wealth of data and possibly to see practical applications, whereas at the same time it is sufficiently self-contained to be realistically mastered without human supervision.

\section{Methods}
\label{sec:method}

Following our informal exposition in Section~\ref{sec:intro}, we distinguish three components, which are learned separately in a sequence: learning a language-independent grounding domain, learning noisy denotation from search logs and finally learning functional composition. While only the first model (prod2vec) is completely unsupervised, it is important to remember that the other learning procedures are only weakly supervised, as the labelling is obtained by exploiting an existing user-machine dynamics to provide noisy labels (i.e. no human labeling was necessary at any stage of the training process).

\textbf{Learning a representation space}.
We train product representation to provide a ``dense ontology'' for the (small) world we want our language to describe. Those representations are known in product search as product embeddings~\citep{Grbovic15}:~\textit{prod2vec} models are word2vec models in which words in a sentence are replaced by products in a shopping session. For this study, we pick CBOW~\citep{DBLP:journals/corr/abs-1804-00306} as our training algorithm, and select $d=24$ as vector size, optimizing hyperparameters as recommended by~\citet{AuthorsKDD2020}; similar to what happens with word2vec, related products (e.g. two pairs of sneakers) end up closer in the embedding space. In the overall picture, the product space just constitutes a grounding domain, and re-using tried and tested~\cite{10.1145/3383313.3411477} neural representations is an advantage of the proposed semantics.

\textbf{Learning lexical denotation}. We interpret clicks on products in the search result page, after a query is issued, as a noisy ``pointing'' signal~\cite{Tagliabue2019LexicalLA}, i.e., a map between text (``shoes'') and the target domain (a portion of the product space). In other words, our approach can be seen as a neural generalization of model-theoretic semantics, where the extension of ``shoes'' is not a discrete set of objects, but a region in the grounding space. Given a list of products clicked by shoppers after queries, we represent meaning through an order-invariant operation over product embeddings (average pooling weighted by empirical frequencies, similar to~\citet{CoveoECNLP20}); following~\citet{Cotter2018InterpretableSF}, we refer to this representation as a~\textit{DeepSet}. Since words are now grounded in a dense domain, set-theoretic functions for NPs~\cite{10.5555/335289} need to be replaced with matrix composition, as we explain in the ensuing section.

\textbf{Learning functional composition.}
Our functional composition will come from the composition of DeepSet representations, where we want to learn a function $f$ : $DeepSet \times DeepSet \rightarrow DeepSet$. We address functional composition by means of two models from the relevant literature~\citep{hartung2017learning}: one, \textit{Additive Compositional Model} (\textbf{ADM}), sums vectors together to build the final DeepSet representation. The second model is instead a \textit{Matrix Compositional Model} (\textbf{MDM}): given in input two DeepSets (for example, one for ``Nike'' and one for ``shoes'') the function we learn as the form $Mv + Nu$, where the interaction between the two vectors is mediated through the learning of two matrices, $M$ and $N$. 
Since the output of these processes is always a DeepSet, both models can be recursively composed, given the form of the function $f$.

\section{Experiments}
\label{sec:experiments}

\begin{table*}[h]
\centering
\small
\begin{tabular}{cccccccc}  \toprule
       \textit{LOBO} & \textbf{$ADM_{p}$} & \textbf{$MDM_{p}$} & \textbf{$ADM_{v}$} & \textbf{$MDM_{v}$} & \textbf{UM} & \textbf{W2V} \\ \midrule
       nDCG &  \underline{0.1821}  &  \textbf{0.2993}  & 0.1635 & 0.0240  & 0.0024 &  0.0098  \\
       Jaccard & \underline{0.0713}   &  \textbf{0.1175} & 0.0450 & 0.0085 & 0.0009 & 0.0052 \\ \bottomrule
\end{tabular}
\caption{Results on \textbf{LOBO} (\textbf{bold} are best, \underline{underline} second best).}
\label{tab:lobo}
\end{table*}

\begin{table*}[h]
\centering
\small
\begin{tabular}{lccccccc} \toprule
\textit{ZT} & \textbf{$ADM_{p}$} & \textbf{$MDM_{p}$} & \textbf{$ADM_{v}$} & \textbf{$MDM_{v}$} & \textbf{UM} & \textbf{W2V} \\ \midrule
\textbf{BAS} (\textit{brand + activity + sortal})\\ \midrule
nDCG & \underline{0.0810}  & \textbf{0.0988} & 0.0600 & 0.0603 & 0.0312 & 0.0064 \\
Jaccard   & \underline{0.0348} &  \textbf{0.0383} & 0.0203 & 0.0214 & 0.0113 & 0.0023 \\ \midrule
\textbf{GAS} (\textit{gender  +  activity  +  sortal})\\ \midrule
nDCG    & \textbf{0.0221} &  0.0078 & 0.0097 & 0.0160 & \underline{0.0190} & 0.0005  \\
Jaccard   & \textbf{0.0083} &  0.0022 & 0.0029 & \underline{0.0056} & 0.0052 & 0.0001 \\ \midrule
\textbf{BGAS} (\textit{brand + gender + activity + sortal})\\ \midrule
nDCG    & \underline{0.0332}  &  \textbf{0.0375} & 0.0118 & 0.0177  &  0.0124 & 0.0059  \\
Jaccard   & \underline{0.0162} & \textbf{0.0163} & 0.0042 & 0.0061 & 0.0044 & 0.0019 \\ \bottomrule
\end{tabular}
\caption{Results on \textbf{ZT} (\textbf{bold} are best, \underline{underline} second best).}
\label{tab:zt}
\end{table*}

\paragraph{Data.} 
We obtained catalog data, search logs and detailed behavioral data (anonymized product interactions) from a partnering online shop,~\textit{Shop X}.~\textit{Shop X} is a mid-size Italian website in the sport apparel vertical\footnote{For convenience of exposition, all queries and examples cited in the paper are translated into English.}. Browsing and search data are sampled from one season (to keep the underlying catalog consistent), resulting in a total of $26,057$ distinct product embeddings, trained on more than $700,000$ anonymous shopping sessions. To prepare the final dataset, we start from comparable literature~\cite{baroni-zamparelli-2010-nouns} and the analysis of linguistic and browsing behavior in~\textit{Shop X}, and finally distill a set of NP queries for our compositional setting. 

In particular, we build a rich, but tractable set by excluding queries that are too rare (<5 counts), queries with less than three different products clicked, and queries for which no existing product embedding is present. Afterwards, we zoom into NP-like constructions, by inspecting which features are frequently used in the query log (e.g. shoppers search for sport, not colors), and matching logs and NPs to produce the final set. Based on our experience with dozens of successful deployments in the space, NPs constitute the vast majority of queries in product search: thus, even if our intent is mainly theoretical, we highlight that the chosen types overlap significantly with real-world frequencies in the relevant domain. Due to the power-law distribution of queries, one-word queries are the majority of the dataset (60\%); to compensate for sparsity we perform data augmentation for rare compositional queries (e.g. ``Nike running shoes''): after we send a query to the existing search engine to get a result set, we simulate $n=500$ clicks by drawing products from the set with probability proportional to their overall popularity~\cite{coveoNAACLINDUSTRY21}\footnote{Since the only objects users can click on are those returned by the search box, query representation may in theory be biased by the idiosyncrasies of the engine. In practice, we confirmed that the embedding quality is stable even when a sophisticated engine is replaced by simple Boolean queries over TF-IDF vectors, suggesting that any bias of this sort is likely to be very small and not important for the quality of the compositional semantics.}. 

The final dataset consists of 104 ``activity + sortal'' \footnote{``Sortal'' refers to a \textit{type} of object: \textit{shoes} and \textit{polo} are sortals, while \textit{black} and \textit{Nike} are not; ``activity'' is the sport activity for a product, e.g. \textit{tennis} for a racket.}  queries -- ``running shoes'' --; 818 ``brand + sortal'' queries -- ``Nike shoes'' --,  and 47 ``gender + sortal'' queries -- ``women shoes''; our testing data consists of 521 ``brand + activity + sortal'' (BAS) triples, 157 ``gender + activity + sortal'' (GAS) triples, 406 ``brand + gender + activity + sortal'' (BGAS) quadruples.\footnote{Dataset size for our compositional tests is in line with intra-textual studies on compositionality~\citep{baroni-zamparelli-2010-nouns,rubinstein-etal-2015-well}; moreover, the lexical atoms in our study reflect a real-world distribution that is independently generated, and not frequency on general English corpora.}

\paragraph{Tasks and Metrics.}
Our evaluation metrics are meant to compare the~\textit{real} semantic representation of composed queries (``Nike shoes'') with the one predicted by the tested models: in the case of the proposed semantics, that means evaluating how it predicts the DeepSet representation of ``Nike shoes'', given the representation of ``shoes'' and ``Nike''. Comparing target vs predicted representations is achieved by looking at the nearest neighbors of the predicted DeepSet, as intuitively complex queries behave as expected only if the two representations share many neighbors. For this reason, quantitative evaluation is performed using two well-known ranking metrics:~\textit{nDCG} and~\textit{Jaccard}~\citep{Vasile2016MetaProd2VecPE,doi:10.1111/j.1469-8137.1912.tb05611.x}. We focus on two tasks: leave-one-brand-out (\textbf{LOBO}) and zero-shot (\textbf{ZT}). In \textbf{LOBO}, we train models over the ``brand + sortal'' queries but we exclude from training a specific brand (e.g., ``Nike''); in the test phase, we ask the models to predict the DeepSet for a seen sortal and an unseen brand. For \textbf{ZT} we train models over queries with two terms (``brand + sortal'', ``activity + sortal'' and ``gender + sortal'') and see how well our semantics generalizes to compositions like ``brand + activity + sortal''; the complex queries that we used at test time are new and unseen. 

\paragraph{Models.} 
We benchmark our semantics (tagged as $p$ in the results table) based on~\textbf{ADM} and~\textbf{MDM} against three baselines: one is another grounded model, where prod2vec embeddings are replaced by image embeddings (tagged as $v$ in the results table), to test the representational capabilities of the chosen domain against a well-understood modality -- image vectors are extracted with ResNet-18, taking the average pooling of the last layer to obtain 512-dimensional vectors; two are intra-textual models, where word embeddings are obtained from state-of-the-art distributional models, BERT (\textbf{UM}) (the Umberto model\footnote{\url{https://huggingface.co/Musixmatch/umberto-commoncrawl-cased-v1}})
and Word2Vec (\textbf{W2V}), trained on textual metadata from \textit{Shop X} catalog. For \textbf{UM}, we extract the 768 dimensional representation from the [CLS] embedding of the 12th layer of the query and learn a linear projection to the product-space (essentially, training to predict the DeepSet representation from text). The generalization to different and longer queries for \textbf{UM} comes from the embeddings of the queries themselves.  Instead, for \textbf{W2V}, we learn a compositional function that concatenates the two input DeepSets, projects them to 24 dimensions, pass them through a Rectified Linear Unit, and finally project them to the product space.\footnote{First results with the same structure as ADM and MDM showed very low performances, thus we made the architecture more complex and non-linear.} We run every model 15 times and report average results; RMSProp is the chosen optimizer, with a batch size of 200, 20\% of the training set as validation set and early stopping with $patience=10$.

\begin{figure*}[!ht]
\centering
    \includegraphics[width=0.7\textwidth]{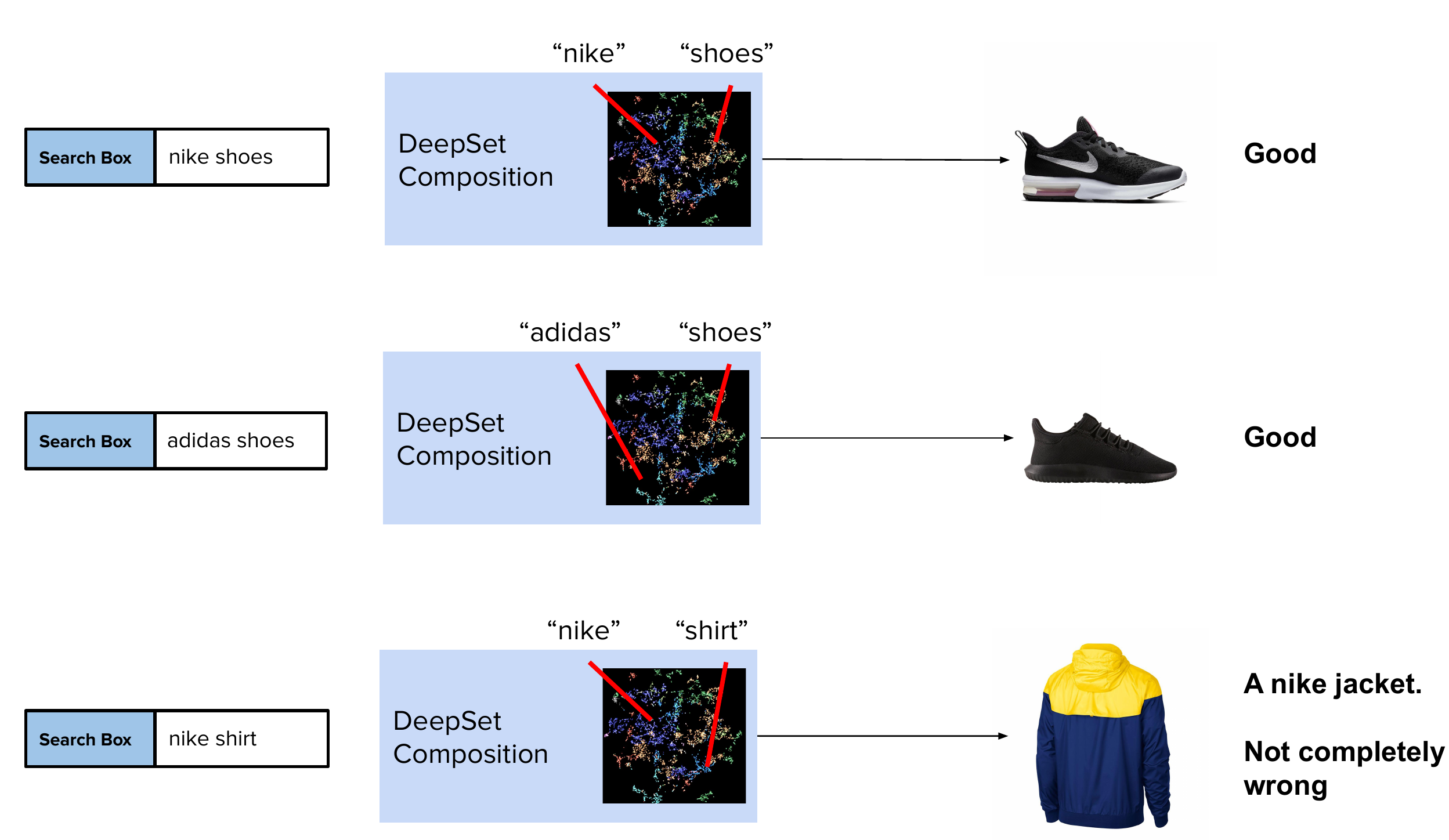}
    \caption{Examples of qualitative predictions made by MDM on the LOBO task.}
    \label{fig:cfm}
\end{figure*}{}

\paragraph{Results.} 
Table~\ref{tab:lobo} shows the results on~\textbf{LOBO}, with grounded models outperforming intra-textual ones, and prod2vec semantics (tagged as $p$) beating all baselines. Table~\ref{tab:zt} reports performance for different complex query types in the zero-shot inference task: grounded models are superior, with the proposed model outperforming baselines across all types of queries. 

\textbf{MDM} typically outperforms \textbf{ADM} as a composition method, except for~\textit{GAS}, where all models suffer from \textit{gender} sparsity; in that case, the best model is~\textbf{ADM}, i.e. the one without an implicit bias from the training. In general, grounded models outperform intra-textual models, often by a wide margin, and prod2vec-based semantics outperforms image-based semantics, proving that the chosen latent grounding domain supports rich representational capabilities. The quantitative evaluations were confirmed by manually inspecting nearest neighbors for predicted DeepSets in the~\textbf{LOBO} setting -- as an example,~\textbf{MDM} predicts for ``Nike shoes'' a DeepSet that has (correctly) all \textit{shoes} as neighbors in the space, while, for the same query, \textbf{UM} suggests \textit{shorts} as the answer. Figure~\ref{fig:cfm} shows some examples of compositions obtained by the MDM model on the LOBO task; the last example shows that the model, given in input the query ``Nike shirt'', does not reply with a shirt, but with a Nike jacket: even if the correct meaning of ``shirt'' was not exactly captured in this contest, the model ability to identify a similar item is remarkable.

\section{Conclusions and Future Work}
\label{sec:conclusion}

In the spirit of~\citet{Bisk2020ExperienceGL}, we argued for grounding linguistic meaning in artificial systems through experience. In our implementation, all the important pieces -- domain, denotation, composition -- are learned from behavioral data. By grounding meaning in (a representation of) objects and their properties, the proposed noun phrase semantics can be learned ``bottom-up'' like distributional models, but can generalize to unseen examples, like traditional symbolic models: the implicit, dense structure of the domain (e.g. the relative position in the space of~\textit{Nike} products and~\textit{shoes}) underpins the explicit, discrete structure of queries picking objects in that domain (e.g. ``Nike shoes'') -- in other words, compositionality is an emergent phenomenon. While encouraging, our results are still preliminary: first, we plan on extending our semantics, starting with Boolean operators (e.g. ``shoes \textit{NOT} Nike''); second, we plan to improve our representational capabilities, either through symbolic knowledge or more discerning embedding strategies; third, we wish to explore transformer-based architectures~\citep{pmlr-v97-lee19d} as an alternative way to produce set-like representations.

We conceived our work as a testable application of a broader methodological stance, loosely following the agenda of the child-as-hacker~\citep{PMID:33012688} and child-as-scientist~\citep{Gopnikarticle} programs. 
Our ``search-engine-as-a-child'' metaphor may encourage the use of abundant real-world search logs to test computational hypotheses about language learning inspired by cognitive sciences~\citep{Carey1978AcquiringAS}.

\section*{Acknowledgments}
We wish to thank Christine Yu, Patrick John Chia and the anonymous reviewers for useful comments on a previous draft. Federico Bianchi is a member of the Bocconi Institute for Data Science and Analytics (BIDSA) and the Data and Marketing Insights (DMI) unit. 

\section*{Ethical Considerations}
User data has been collected in the process of providing business services to the clients of~\textit{Coveo}: user data is collected and processed in an anonymized fashion, in full compliance with existing legislation (GDPR). In particular, the target dataset uses only anonymous uuids to label sessions and, as such, it does not contain any information that can be linked to individuals.

\bibliography{anthology,custom}
\bibliographystyle{acl_natbib}

\end{document}